\documentclass{article}
\usepackage{spconf,amsmath,graphicx,hyperref}
\usepackage{subcaption}
\usepackage[export]{adjustbox}
\usepackage{mathtools}
\usepackage{booktabs}
\usepackage{siunitx}
\usepackage{amssymb}

\title{SimulSense: Sense-Driven Interpreting for Efficient \\Simultaneous Speech Translation}

\name{Haotian Tan \qquad Hiroki Ouchi \qquad Sakriani Sakti}
\address{Nara Institute of Science and Technology, Japan}
\begin{document}
\ninept
\maketitle

\begin{abstract}
How to make human-interpreter-like read/write decisions for simultaneous speech translation (SimulST) systems?
Current state-of-the-art systems formulate SimulST as a multi-turn dialogue task, requiring specialized interleaved training data and relying on computationally expensive large language model (LLM) inference for decision-making. 
In this paper, we propose SimulSense, a novel framework for SimulST that mimics human interpreters by continuously reading input speech and triggering write decisions to produce translation when a new sense unit is perceived.
Experiments against two state-of-the-art baseline systems demonstrate that our proposed method achieves a superior quality-latency tradeoff and substantially improved real-time efficiency, where its decision-making is up to \textbf{9.6$\times$} faster than the baselines.
\end{abstract}

Code is available at \url{https://github.com/Tan-ht/SimulSense-ST}.

\section{Introduction}
\label{sec:intro}

Simultaneous speech translation (SimulST) is a challenging task to perform translation in real-time with low latency while maintaining high translation quality.
A key component of such systems is the decision policy, which determines whether the currently available source context is sufficient for accurate translation or how many predicted tokens can be reliably displayed to users.

Early works on the SimulST have primarily relied on two paradigms: (1) designing dedicated training strategies to learn data-driven decision policies~\cite{ma2020simulmt, zeng2021realtrans}, or (2) adapting pre-trained offline ST models with hand-crafted, rule-based decision policies~\cite{wait-k, lowlatency, alignatt}. However, these methods either require training multiple models for different latency scenarios or are highly dependent on the design of the rule-based decision policy.
Recently, SimulST has been framed as a multi-turn dialogue task using Large Language Models (LLMs), where speech segments are treated as user prompts and translations as assistant responses~\cite{wang-etal-2025-conversational, fu2025efficient, ouyang-etal-2025-infinisst}. 
To control the quality-latency trade-off, \cite{ouyang-etal-2025-infinisst} employs a simple fixed decision policy that ensures translation begins only after a fixed number of source tokens have arrived. \cite{fu2025efficient, cheng2024towards} enable the LLM to learn a data-driven read/write policy that determines whether the accumulated source context is sufficient for translation. These multi-turn dialogue systems have achieved state-of-the-art performance in recent years.
However, they require interleaved SimulST data for training, which is challenging to collect due to the high cost of manual annotation. 
\cite{fu2025efficient,cheng2024towards} propose prompting a powerful LLM to generate synthetic SimulST data, aiming to mimic human interpreter behavior and learn an optimal read/write policy. Nevertheless, their performance is highly dependent on the LLM capabilities and prompt engineering. \cite{ouyang-etal-2025-infinisst} constructs training data from offline ST corpus using MFA and SimAlign while its decision policy does not mimic the human interpreters, making it sub-optimal.

\begin{figure}[t]
    \centering
    \includegraphics[width=0.8\columnwidth]{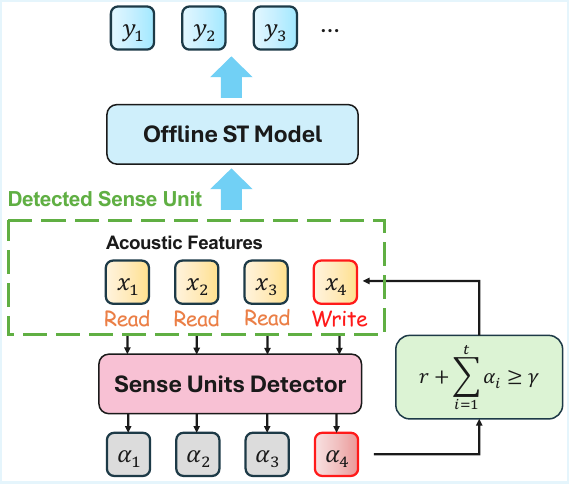}
    \caption{The Sense Units Detector (SUD) assigns a weight to each acoustic feature of the input stream. When the accumulated weight sum (including a prior residual weight $r$) exceeds a triggering threshold $\gamma$, a sense unit is detected, which triggers the offline ST model to produce translations.}
    \label{fig:1}
\end{figure}

\begin{figure*}[ht]
  \centering
  \includegraphics[
    width=0.95\linewidth,
    max height=0.75\textheight,
    keepaspectratio
  ]{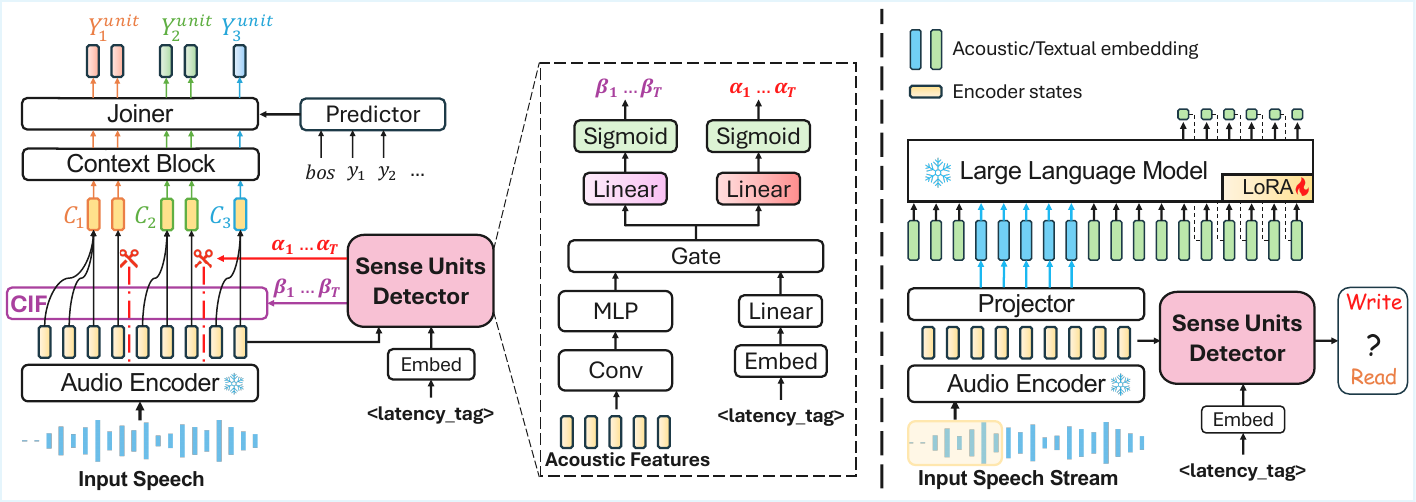}
  \caption{The overview of our SimulSense framework. \textbf{Left:}A Sense-Aware Transducer (SAT) training pipeline that guides a lightweight Sense Units Detector (SUD) model in learning human-interpreter-like sense unit segmentation. \textbf{Right:}The SUD model dynamically perceives sense units from acoustic features and drives the LLM-based offline ST model to perform simultaneous translation.}
  \label{fig:framework}
\end{figure*}

In this paper, we propose SimulSense, a sense-aware framework for SimulST to achieve a better quality-latency trade-off. 
SimulSense is motivated by the cognitive process of human interpreters, where they attentively listen to the speaker's speech and begin translation as soon as they perceive a sense unit (semantic chunk), which is the smallest linguistic unit capable of conveying a complete thought independently~\cite{lederer1978simultaneous, jones2014conference}. This behavior enables low-latency, cognitively efficient, and semantically coherent output.
To mimic this behavior, as illustrated in Figure \ref{fig:1}, our approach continuously processes the input speech until a sense unit is detected by a dedicated Sense Units Detector (SUD) model. Upon detection, the write decision will be triggered to send the acoustic features to an offline ST model to produce translation.
Despite~\cite{fu2025efficient,cheng2024towards} also aiming to detect a sense unit to initiate translation, they require a full forward pass through the LLM to make decisions, which is inefficient even with the aid of the KV cache mechanism. In contrast, our method employs a lightweight SUD model that makes decisions directly from the audio encoder output, enabling efficient, real-time decision-making without costly LLM inference. Furthermore, prior works rely heavily on specialized interleaved training data, which may compromise translation quality. Our method instead leverages a well-established offline ST model trained on ground-truth parallel data, ensuring robust and high-quality translation capability. Our contributions are summarized as follows:
\begin{itemize}
    \item We propose SimulSense, a novel framework that mimics human interpreters to efficiently drive the simultaneous speech translation with dynamically detected sense units.
    \item We design a Sense Units Detection mechanism as well as its training pipeline, avoiding costly LLM inference and enabling efficient read/write decisions for SimulST.
\end{itemize}

\section{Related Work}

\subsection{Continuous Integrate-and-Fire for SimulST}
Continuous Integrate-and-Fire (CIF)~\cite{cif} is a soft and monotonic alignment mechanism originally proposed for ASR to learn the acoustic boundaries. 
The CIF operates in two modes: integration and firing. The integration mode uses a group of weights for speech frames to dynamically accumulate information until the sum of weights exceeds a certain threshold, at which point it enters the firing mode to produce an output for the accumulated information.
In recent years, CIF has gained attention in the speech translation community. \cite{dong-etal-2022-learning} integrated CIF into SimulST to locate acoustic boundaries, while \cite{chang22f_interspeech,deng2024label} employed it to decide how much input to read per translation token. Additionally, \cite{deng-etal-2025-simuls2s} used CIF to locate word boundaries to bring SimulST closer to text-to-text translation.
Our work differs from these approaches. Instead of identifying acoustic or word boundaries, we aim to accumulate acoustic information corresponding to a semantically independent sense unit. Furthermore, we only use CIF to guide the learning of the Sense Units Detector in an ASR training pipeline. In our SimulST system, we only accumulate information without engaging in the integration and firing process.

\subsection{Adapting Offline Systems for SimulST}
Offline ST models consume the full source speech to achieve superior translation quality. Early works have explored leveraging offline models for SimulST through fixed, rule-based decision policies. \cite{wait-k} waits for k speech segments before the translation starts. \cite{lowlatency} detects reliable model predictions to display to users by making an agreement between two consecutive chunks. \cite{alignatt} exploit the cross-attention mechanism to make such decisions.
Unlike these fixed-policy approaches, we aim to develop a flexible and adaptive decision policy that mimics human interpreter behavior.

\section{Proposed Method}
Figure~\ref{fig:framework} shows the overview of our SimulSense framework, which comprises a Sense Units Detector (SUD) for identifying semantic boundaries, a sense-aware transducer (SAT) training pipeline for optimizing read/write decisions, and an offline-trained speech translation model for high-quality translation generation.

\subsection{Sense Units Data Creation}
\label{sec:2.1}
Inspired by recent work~\cite{cheng2024towards} that leverages LLMs for SimulST data creation, we create sense units data from existing speech-to-text datasets using a Qwen3-32B model. Given a speech-to-text sample $(S,Y)$, where $S$ is the speech sequence and $Y$ is the source transcription, we prompt the LLM to segment $Y$ into $N$ sense units $Y_{units}=(Y^{unit}_1,...Y^{unit}_N)$.
Following~\cite{fu2025efficient}, we use three latency settings (low, medium, high) that result in different segmentation densities: $N_{low}>N_{medium}>N_{high}$. Each sample is accompanied by a latency tag ($\text{\textless low\textgreater, \textless medium\textgreater, or \textless high\textgreater}$) to enable latency-aware training.

\subsection{Sense-Aware Transducer (SAT)}
As illustrated in the left panel of Figure \ref{fig:framework}, the Sense-Aware Transducer (SAT) guides the SUD by integrating the CIF mechanism into a transducer-based ASR training pipeline.
The SUD model functions as a weight predictor in the CIF mechanism. Unlike conventional CIF weight predictors, the SUD takes encoder output acoustic features $H = (h_1, \ldots, h_T)$ and latency tag embeddings as inputs to predict two weight groups: $\mathrm{A} = (\alpha_1, \ldots, \alpha_T)$ and $\mathrm{B} = (\beta_1, \ldots, \beta_T)$ for each acoustic feature $h_i$, where $T$ is the total number of frames.
We perform CIF-like continuous weight accumulation using weights group $\mathrm{A}$ and dynamically segment the acoustic feature sequence into $K$ non-overlapping segments $\{\mathcal{U}_k\}_{k=1}^K$ by applying a threshold to the cumulative $\alpha$-weights. 
Specifically, a segmentation boundary is triggered at frame $t$ if the cumulative weights since the last boundary exceed a triggering threshold $\gamma$, which is fixed to 1.0 during training:
\begin{equation}
    r + \sum_{i = j + 1}^{t} \alpha_i \geq \gamma,
    \label{eq:1}
\end{equation}
where $j$ denotes the frame index of the previous boundary, and $r \in [0, \gamma)$ is the residual weight from the prior segment. 
Upon triggering a boundary at frame $t$, the residual weight is updated as:
\begin{equation}
    r \leftarrow r + \sum_{i = j + 1}^{t} \alpha_i - \gamma,
\end{equation}
This ensures that excess weight beyond the threshold is preserved for the next segment.

During training, given the sense units data $(S,Y_{units})$ containing $N$ sense units, we apply the standard CIF scaling strategy to the weights group $\mathrm{A}$ to ensure:
\begin{equation}
    \sum^{T}_{t=1}\alpha_t = N-1.
\end{equation}
This constraint guarantees exactly $N-1$ boundary triggers, producing $N$ acoustic feature segments $\{\mathcal{U}_k\}_{k=1}^N$ that correspond to the $N$ sense units.
We then generate unit-level integrated acoustic features by applying the CIF mechanism independently to each sense unit feature $\mathcal{U}_k$ using weights group $\mathrm{B}$ and a fixed firing threshold $\lambda$:
\begin{equation}
    C_k = \mathrm{CIF}\left( \mathcal{U}_k; \, \mathrm{B}, \, \lambda = 1.0 \right),
\end{equation}
where $C_k = (c_k^1, \dots, c_k^{M_k})$ represents the CIF features for the $k$-th sense unit and ${M_k}$ denotes the feature length.
Given the target transcription $Y^{unit}_k = (y_k^1, \dots, y_k^{L_k})$ for the $k$-th unit, where $L_k$ is the number of target tokens, we apply the scaling strategy to weights $\mathrm{B}$ before CIF to enforce ${M_k} = L_k$. This ensures one-to-one alignment between the CIF features $C_k = (c_k^1, \dots, c_k^{M_k})$ and target tokens, such that $c_k^j \leftrightarrow y_k^j$.
Following \cite{cif-t}, we append a context block after the CIF module to enhance contextual dependencies among the CIF outputs.

This process yields length-aligned joiner inputs $(\{C_k\}^N_{k=1}, Y)$ and produces three-dimensional output logits. Unlike traditional RNN-T training, we employ cross-entropy loss $\mathcal{L}_{Joint}$ for more efficient optimization.
To force the model to learn alignment between acoustic feature segments and sense units, we introduce two quantity losses. 
First, we constrain the weights accumulation to the boundary triggers $N-1$ to ensure the total number of acoustic feature segments matches the number of sense units $N$:
\begin{align}
    \mathcal{L}_{Qua1} = |\sum_{t=1}^{T}\alpha_t - (N-1)|.
\end{align}
Second, we ensure that CIF outputs match the target token count for each sense unit:
\begin{align}
    \mathcal{L}_{Qua2} = \sum_{k=1}^{N}{|\sum_{i=1}^{M_k}{\beta_k^i - L_k}|}.
\end{align}
Additionally, we apply cross-entropy loss $\mathcal{L}_{LM}$ to the predictor model to enhance contextual semantic understanding. The total training loss combines all components:
\begin{align}
    \mathcal{L}_{Total} = \mathcal{L}_{Joint} + \mathcal{L}_{Qua1} + \mathcal{L}_{Qua2} + \mathcal{L}_{LM}.
\end{align}

\subsection{Read/Write Decision Policy}
We employ an offline ST model to perform translation, which shares the same pre-trained acoustic encoder as the SAT network. During inference, as illustrated in the right panel of Figure \ref{fig:framework}, we construct the SimulST pipeline by integrating the pre-trained SUD model with the offline ST model.
The audio encoder processes the input speech stream in a chunk-by-chunk manner. Given a latency tag, the SUD model dynamically weights the acoustic features and triggers the offline ST model to generate translation when a new sense unit is detected (i.e., when Equation~\ref{eq:1} is satisfied).

\begin{figure*}[htbp]
    \centering
    \begin{subfigure}[b]{0.33\linewidth}
        \includegraphics[width=\linewidth]{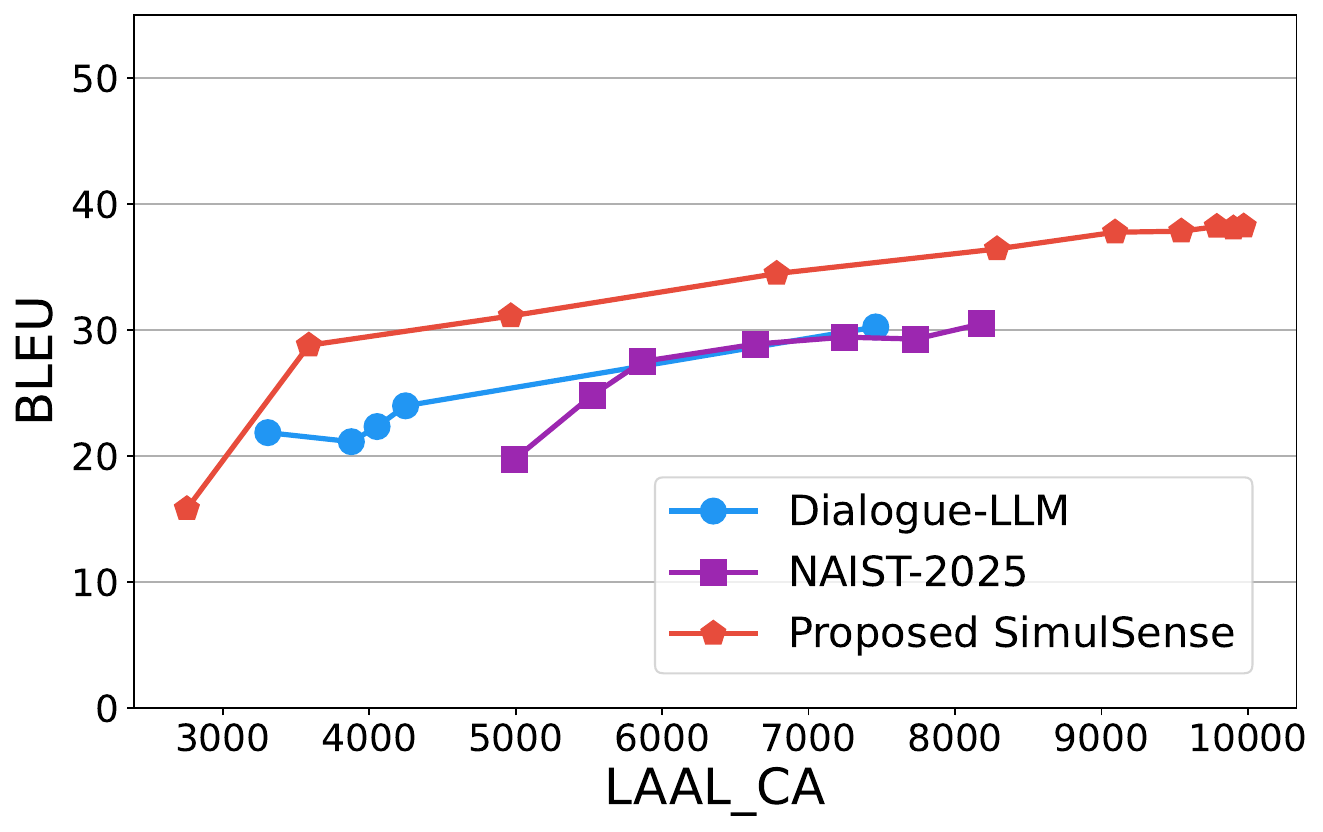}
        \caption{En-De}
        \label{fig:1a}
    \end{subfigure}
    \hfill 
    \begin{subfigure}[b]{0.33\linewidth}
        \includegraphics[width=\linewidth]{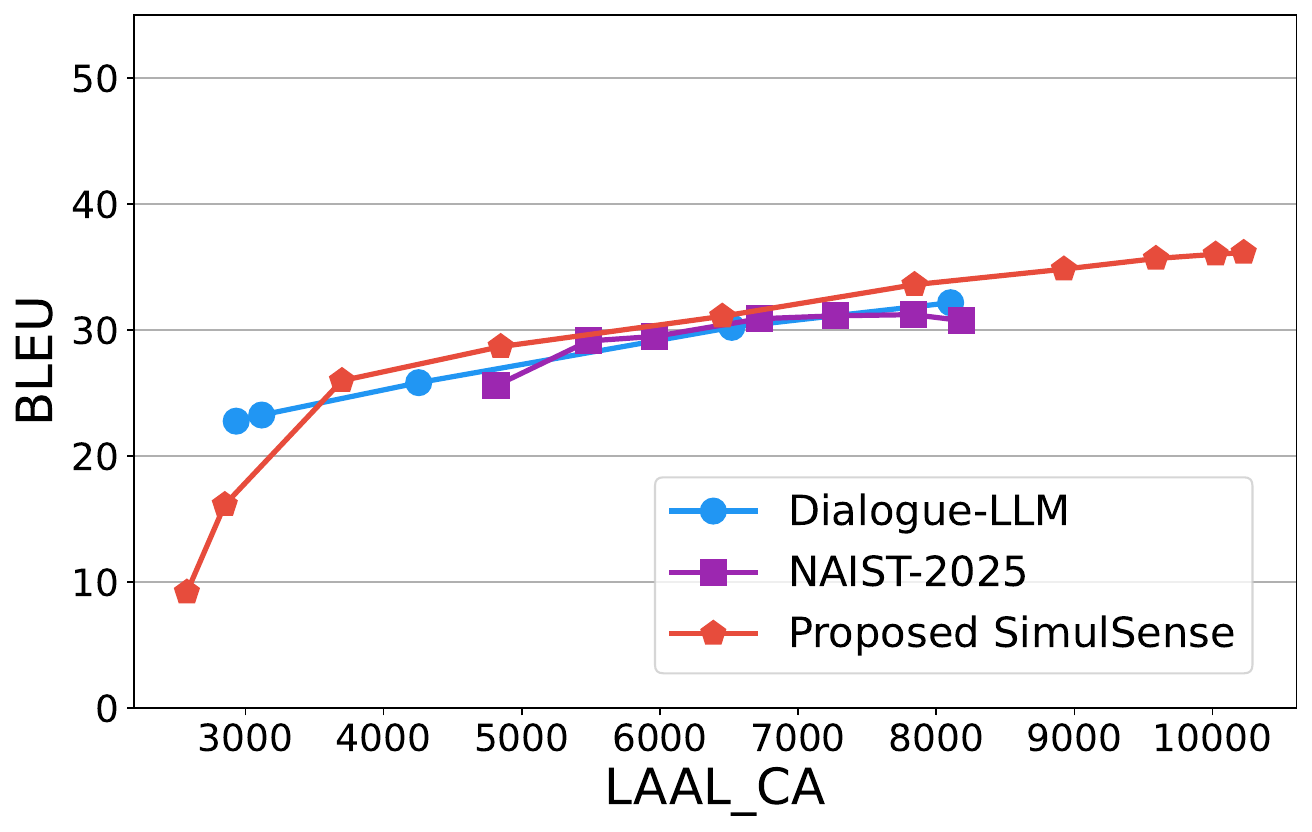}
        \caption{En-Ja}
        \label{fig:1b}
    \end{subfigure}
    \hfill
    \begin{subfigure}[b]{0.33\linewidth}
        \includegraphics[width=\linewidth]{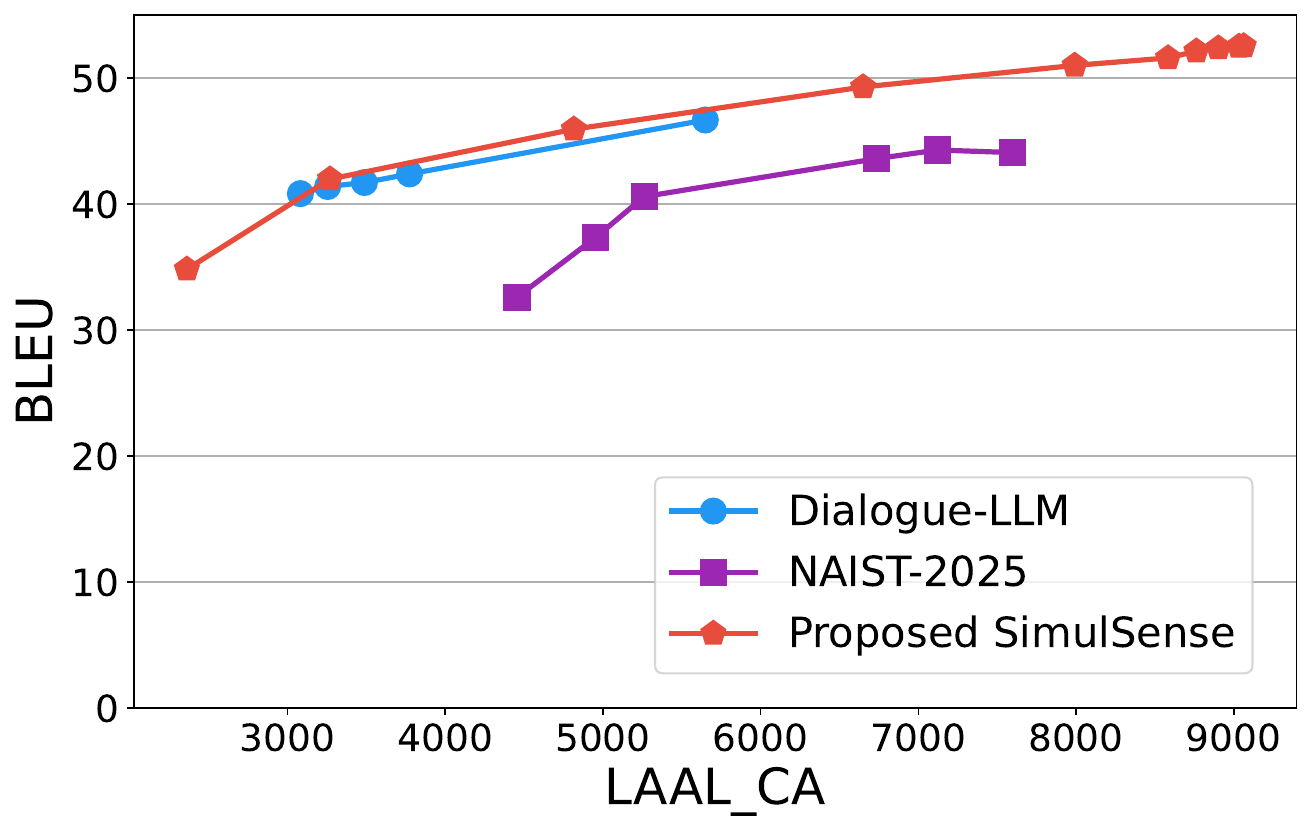}
        \caption{En-Zh}
        \label{fig:1c}
    \end{subfigure}
    \caption{Quality-latency trade-off (latency in ms) compared to two state-of-the-art SimulST systems on the En$\rightarrow$De, En$\rightarrow$Ja, and En$\rightarrow$Zh directions of the ACL 60/60 test set. SimulSense achieves a superior quality-latency tradeoff across all language pairs.}
    \label{fig:tradeoff}
\end{figure*}

\section{Experimental Setup}

\subsection{Data}
The International Conference on Spoken Language Translation (IWSLT) is an annual conference and evaluation campaign that promotes research on spoken language translation. In this paper, we follow the data condition of the IWSLT 2025 Simultaneous Track \cite{agostinelli-etal-2025-findings} to use CoVoST-2 \cite{wang-etal-2020-covost} as the training set and ACL 60/60 \cite{salesky-etal-2023-evaluating} for both validation and evaluation.
We conduct experiments on three translation directions: English-to-German (En$\rightarrow$De), English-to-Japanese (En$\rightarrow$Ja), and English-to-Chinese (En$\rightarrow$Zh).

\subsection{Model Architecture}
We use Whisper-large-v3~\cite{pmlr-v202-radford23a} as the audio encoder.
In the SAT network, a Conformer encoder serves as the context block, followed by a UGBP joiner~\cite{ugbp}. The predictor is Qwen-3-0.6B, fine-tuned with LoRA during SAT training.
The SUD model comprises a two-layer convolution block, a two-layer MLP, and a gating mechanism.
Our offline ST model stacks the Whisper encoder, a two-layer linear projector, and the Qwen-3-8B LLM. During training, we freeze the Whisper encoder and finetune the LLM with LoRA.

\subsection{Baselines}
We compare our proposed method against two state-of-the-art SimulST systems.

\noindent\textbf{NAIST-2025~\cite{tan-etal-2025-naist}}: This model adapts an LLM-based offline ST system for SimulST by applying a Local Agreement (LA) policy~\cite{lowlatency}. At IWSLT 2025~\cite{agostinelli-etal-2025-findings}, it was ranked 1st for the En$\rightarrow$Ja language pair and 2nd for both En$\rightarrow$Zh and En$\rightarrow$De in the low-latency setting.

\noindent\textbf{Dialogue-LLM}: This baseline is built upon the methodology described in~\cite{fu2025efficient}, utilizing the multi-turn dialogue mechanism of an LLM. To ensure a fair comparison, we use the same audio encoder, projector, LLM, and CovoST-2-derived training data as our SimulSense framework. We show later that the Dialogue-LLM baseline outperforms the NAIST-2025 system in terms of quality-latency tradeoff in Section \ref{sec:tradeoff}.

\subsection{Evaluation Setup}
We use BLEU ($\uparrow$) to evaluate the translation quality. Latency is measured using the LAAL metric~\cite{laal}. We report the computation-aware latency with all evaluations performed on an NVIDIA RTX A6000 GPU. 
To further assess the efficiency of the read/write decision policy across systems, we report the average inference time per decision and the Real-Time Factor (RTF).
We use different triggering thresholds $\gamma=\{0.5,1.0,\dots,5.0\}$ to control the quality-latency tradeoff of our proposed method.

\section{Results}

\subsection{Quality-Latency Tradeoff}
\label{sec:tradeoff}
Figure \ref{fig:tradeoff} illustrates the computation-aware quality-latency tradeoff of our SimulSense framework against the baselines. Our system significantly outperforms both baselines for the En$\rightarrow$De language pair, achieving a maximum BLEU improvement of 6.9 against Dialogue-LLM (at around 3.5s latency) and 11.4 against NAIST-2025 (at around 5.0s latency).
Furthermore, it also consistently outperforms both baselines for the En$\rightarrow$Zh direction. For the En$\rightarrow$Ja pair, the translation quality of our system surpasses that of the baselines when latency exceeds 3 seconds. 
Notably, the proposed method is capable of achieving a wider latency span, ranging from approximately 2.5s to 10s, by simply adjusting the SUD triggering threshold. This flexibility makes it highly suitable for real-world deployment.

\subsection{Efficiency on Read/Write Decision Policy}
Unlike NAIST-2025 and Dialogue-LLM, which derive their decisions from LLM outputs, our SimulSense framework employ a more efficient decision module comprising an audio encoder and the SUD model. As shown in Table~\ref{tab:efficiency}, this decision module achieves a significantly lower average inference time. Specifically, it is 3.0$\times$ faster than Dialogue-LLM and 9.6$\times$ faster than NAIST-2025 per decision. Moreover, its RTF is 8.1$\times$ and 11.3$\times$ lower than those of the two baselines, respectively, indicating a substantial improvement in real-time efficiency.
\begin{table}[ht]
\centering
\caption{Efficiency comparison of the decision policies across different systems.}
\vspace{-0.1cm}
\label{tab:efficiency}
\footnotesize
\begin{tabular}{lcc}
\toprule
Model        & Avg. Inference Time (ms) & RTF   \\
\midrule
NAIST-2025   & 371.3               & 0.180 \\
Dialogue-LLM & 116.2               & 0.130 \\
\textbf{Proposed SimulSense}          & \textbf{38.6}       & \textbf{0.016} \\
\bottomrule
\end{tabular}
\end{table}

\vspace{-0.1cm}
\subsection{Impact of Latency Tags}
\label{sec:tag_impact}
\begin{figure}[htbp]
    \centering
    \includegraphics[width=0.65\linewidth]{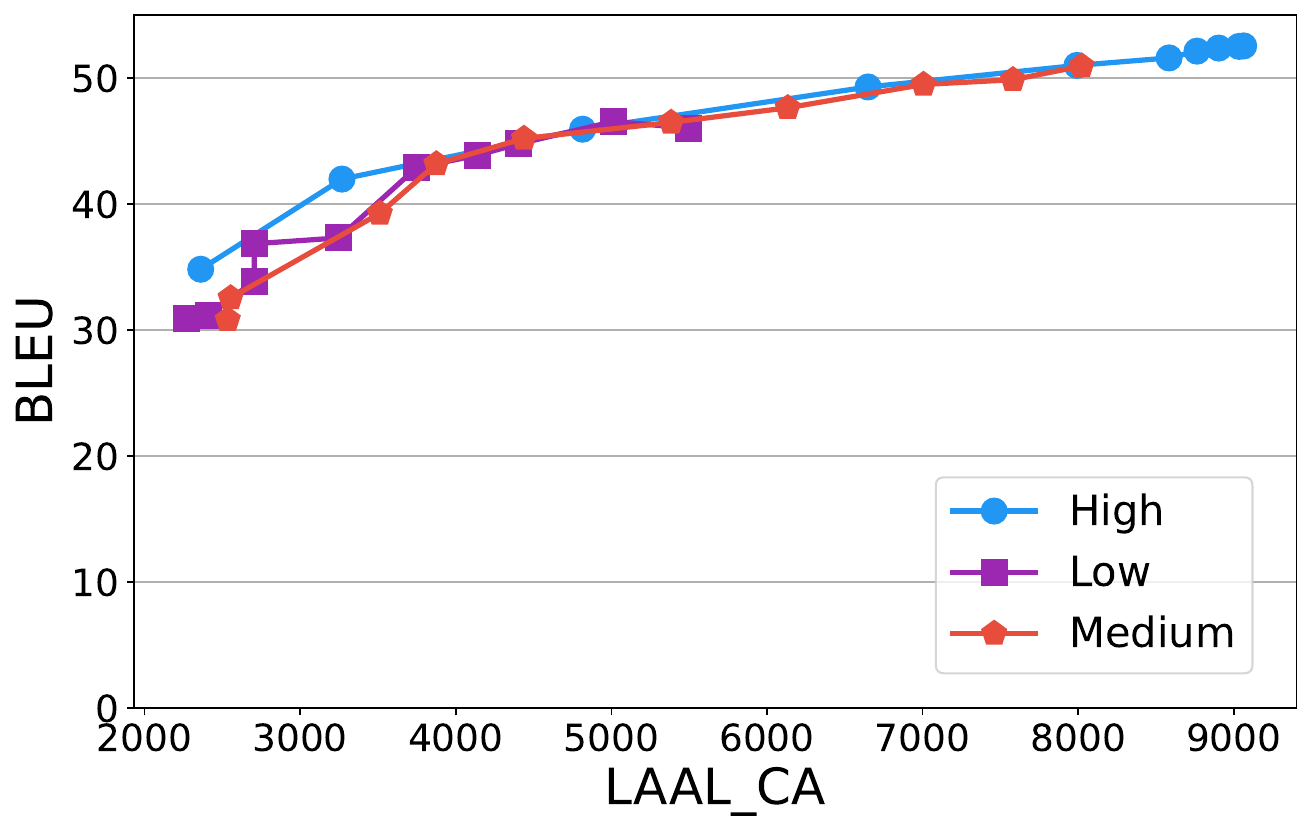}
    \caption{The impact of using different latency tags (latency in ms).}
    \vspace{-0.1cm}
    \label{fig:tag_ablation}
\end{figure}

We also investigated the use of latency tags to control the quality-latency tradeoff. However, as shown in Figure \ref{fig:tag_ablation}, our results suggest that these tags primarily constrain the maximum latency and do not significantly impact the overall tradeoff. Consequently, we used a high-latency tag for all experiments.

\vspace{-0.1cm}
\subsection{WER of Sense-Aware Transducer}
As shown in Table~\ref{tab:model_comparison}, we evaluated three SAT variants with varying numbers of context block layers and attention heads. 
We selected SAT-Small as the final architecture, as it achieved a superior WER of 67.7\% on the ACL 60/60 test set.
Interestingly, although the SAT training yielded a relatively high WER, the SUD model was still able to make effective decisions for the SimulST task. This may be because during SAT training, the SUD model effectively learned to identify sense unit boundaries but did not perform well at transcribing the unit-level acoustic information integrated by the CIF mechanism.
\begin{table}[htbp]
\centering
\caption{WER comparison across SAT variants with varying context block layers and attention heads.}
\vspace{-0.1cm}
\label{tab:model_comparison}
\footnotesize
\setlength{\tabcolsep}{2pt}  
\begin{tabular}{l c c c S[table-format=3.0] S[table-format=2.1]}
\toprule
\textbf{Model} & \textbf{Context Block} & \textbf{Layer} & \textbf{Heads} & {\textbf{\#Params (M)}} & {\textbf{WER (\%)}} \\
\midrule
SAT-Mini  & $\times$     & — & — & 41   & 74.5 \\
SAT-Small & $\checkmark$ & 2 & 2 & 100  & \textbf{67.7} \\
SAT-Large & $\checkmark$ & 8 & 4 & 232  & 69.1 \\
\bottomrule
\end{tabular}
\end{table}

\vspace{-0.1cm}
\section{Conclusion}
We proposed SimulSense, a sense-aware framework that enables SimulST to make human-interpreter-like read/write decisions. Unlike existing methods that rely on LLM inference, SimulSense employs a highly efficient Sense Units Detector (SUD) model for decision-making and leverages an offline-trained speech translation model to conduct high-quality translation. Experiments against two state-of-the-art SimulST systems across three translation directions demonstrate that SimulSense achieves a superior quality-latency tradeoff and substantially improved real-time efficiency.

\section{Acknowledgments}
Part of this work was supported by JSPS KAK-
ENHI Grant Numbers JP21H05054, JP23K21681, and JST NEXUS (JPMJNX25C1).


\bibliographystyle{IEEEbib}

\bibliography{refs,refs2}

\end{document}